\documentclass[conference]{IEEEtran}
\IEEEoverridecommandlockouts
\usepackage{cite}
\usepackage{authblk}
\usepackage{amsmath,amssymb,amsfonts}
\usepackage{algorithmic}
\usepackage[titlenumbered, ruled, linesnumbered, vlined]{algorithm2e}
\usepackage{graphicx}
\usepackage{caption,subcaption}
\usepackage{textcomp}
\usepackage{xcolor, colortbl,}
\usepackage{todonotes,soul}
\usepackage{multirow}
\usepackage{url}
\usepackage{hyperref}
\def\BibTeX{{\rm B\kern-.05em{\sc i\kern-.025em b}\kern-.08em
    T\kern-.1667em\lower.7ex\hbox{E}\kern-.125emX}}

\usepackage{soul}
\soulregister\cite7
\soulregister\ref7
\soulregister\pageref7

\def\fixme#1{\typeout{FIXED in page \thepage : {#1}}
\bgroup \color{red}{[FIXME: {#1}]} \egroup}

\hypersetup{
    colorlinks=true,
    linkcolor=blue,
    urlcolor=blue, 
    citecolor=blue, 
}

\begin{document}

\title{VALO: A Versatile Anytime Framework for LiDAR based Object Detection Deep Neural Networks}
\author[1]{Ahmet Soyyigit}
\author[2]{Shuochao Yao}
\author[3]{Heechul Yun}
\affil[1,3]{University of Kansas, Lawrence, KS\\ \{ahmet.soyyigit,heechul.yun\}@ku.edu}
\affil[2]{George Mason University, Fairfax, VA\\ shuochao@gmu.edu}
\maketitle

\begin{abstract}
This work addresses the challenge of adapting dynamic deadline requirements for LiDAR object detection deep neural networks (DNNs). The computing latency of object detection is critically important to ensure safe and efficient navigation. However, state-of-the-art LiDAR object detection DNNs often exhibit significant latency, hindering their real-time performance on resource-constrained edge platforms. Therefore, a tradeoff between detection accuracy and latency should be dynamically managed at runtime to achieve optimum results. 

In this paper, we introduce VALO (Versatile Anytime algorithm for LiDAR Object detection), a novel data-centric approach that enables anytime computing of 3D LiDAR object detection DNNs. VALO employs a deadline-aware scheduler to selectively process input regions, making execution time and accuracy tradeoffs without architectural modifications. Additionally, it leverages efficient forecasting of past detection results to mitigate possible loss of accuracy due to partial processing of input. Finally, it utilizes a novel input reduction technique within its detection heads to significantly accelerate execution without sacrificing accuracy. 

We implement VALO on state-of-the-art 3D LiDAR object detection networks, namely CenterPoint and VoxelNext, and demonstrate its dynamic adaptability to a wide range of time constraints while achieving higher accuracy than the prior state-of-the-art. 
Code is available at \href{https://github.com/CSL-KU/VALO}{github.com/CSL-KU/VALO}.
\end{abstract}

\begin{IEEEkeywords}
LiDAR, 3D object detection, Anytime computing
\vspace{-10pt}
\end{IEEEkeywords}

\section{Introduction}\label{sec-intro}

Perception plays a vital role in autonomous vehicles. Its primary objective is to identify and categorize objects of interest (e.g., cars, pedestrians) within the operational environment. While humans excel at this task effortlessly, it presents a significant challenge for computers. 
For object detection in three dimensional (3D) space, 
LiDAR based object detection deep neural networks (DNNs)~\cite{centerpoint,pointpillars,voxelnext} have emerged as an effective approach as they can provide highly accurate position, orientation, size, and velocity estimates.

In autonomous vehicles, however, the object detection results must not only be accurate but also timely, as outdated results are of little use in the path planning of a fast-moving autonomous vehicle. Unfortunately, LiDAR object detection DNNs are often computationally expensive and thus exhibit significant latency, especially when running on resource-constrained embedded computing platforms. Moreover, they lack the ability to dynamically trade execution time and accuracy, which makes it difficult to adapt to dynamically changing real-time requirements in
autonomous vehicles~\cite{d3,li2023red}. 
For example, when a vehicle moves at a high speed, fast detection may be more important than high accuracy (e.g., correct object classification) in order to avoid collision in a timely manner. On the other hand, when the vehicle moves slowly in a complex urban environment, accurate detection may be more important than fast detection for safe navigation. 


To enable schedulable trade-offs between accuracy and latency in perception, prior research efforts have focused on vision-based DNNs~\cite{anytimenet, apnet, imprecisecomp, rttasksched, abc}. Model-level innovations such as early-exit architectures~\cite{rttasksched} have been widely adopted, where these models incorporate additional output layers at intermediate stages, allowing the network to make predictions before the full depth of the model is utilized.
Nonetheless, these enhancements come with a trade-off. 
The repeated activation of intermediate
output layers at several phases leads to a significant increase in
computational overhead. 
This issue is particularly pronounced in applications requiring complex detection heads capable of producing granular object-level predictions, such as LiDAR based object detection and segmentation tasks. 
Recently, AnytimeLidar~\cite{alidar} introduced a capability to bypass certain components and detection heads in a LiDAR object detection DNN to enable latency and accuracy trade-offs at runtime.
However, such model-level improvements may not work on different model architectures, which are constantly evolving. 

\begin{figure*}[t]
\centerline{\includegraphics[scale=0.4]{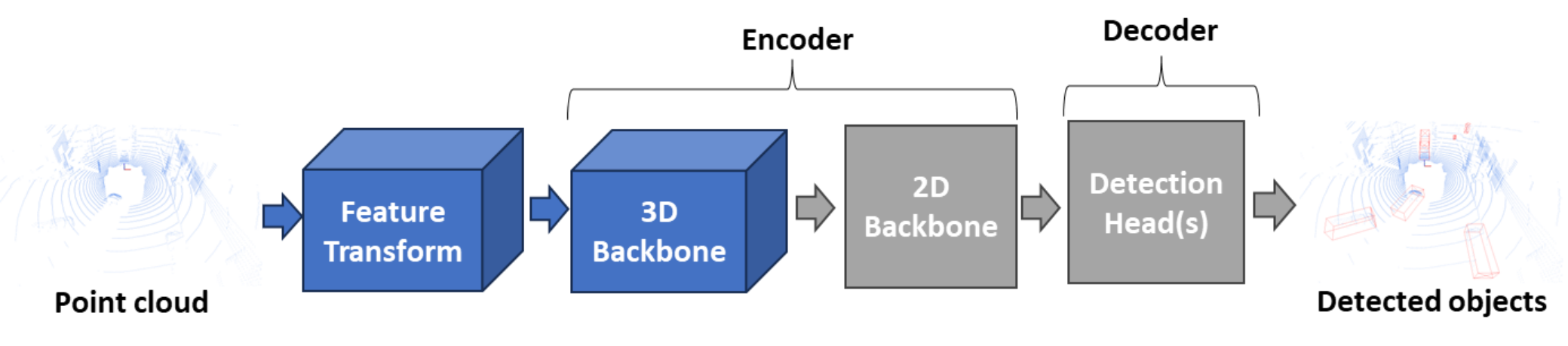}}
\caption{General LiDAR object detection DNN architecture.}
\label{fig:baseline}
\vspace{-15pt}
\end{figure*}

In this work, we present VALO (Versatile Anytime algorithm for LiDAR Object detection), a novel data-centric approach to enable anytime computing in processing LiDAR based object detection DNNs. 
VALO selectively processes subsets of periodically given input data with the aim of maximizing detection accuracy while respecting the deadline constraint. It implements a deadline-aware scheduler that splits the detection area into regions and schedules them to reduce computational costs while considering the accuracy impacts. To minimize potential accuracy loss, VALO employs a lightweight forecasting algorithm to predict the current poses of previously detected objects based on a simple physics model. The forecasted objects are merged with the DNN detected ones through non-maximum suppression to improve overall accuracy. In addition, VALO implements a novel input reduction technique within its detection heads. This technique reduces the input volume to be processed by a factor of ten for the convolutions responsible for delivering object attributes. Importantly, it accomplishes this without any loss in accuracy by eliminating unnecessary computation in the areas where no object prediction exists.

We have implemented VALO on top of two state-of-the-art LiDAR object detection DNNs~\cite{centerpoint,voxelnext} and evaluated them using a large-scale autonomous driving dataset, nuScenes~\cite{nuscenes}. We utilized the Jetson AGX Xavier~\cite{jetson-agx} as the testing platform, a commercially available off-the-shelf embedded computing platform. The results demonstrate that VALO enables the anytime capability across a wide spectrum of timing constraints, 
while achieving higher accuracy across all deadline constraints compared to the baseline lidar object detection DNNs~\cite{centerpoint,voxelnext} and a prior anytime approach~\cite{alidar}.

In summary, we make the following contributions:
\begin{itemize}
\item We propose a novel data scheduling framework for LiDAR object detection DNNs that enables latency and accuracy tradeoffs at runtime.
\item We apply our approach to two state-of-the-art LiDAR object detection DNNs and show its effectiveness and generality on a real platform using a representative autonomous driving dataset. 
\end{itemize}

The rest of the paper is organized as follows: We provide the necessary background in Section~\ref{sec:background} and present motivation in Section~\ref{sec:motivation}. We describe our approach in Section~\ref{sec:VALO} and present the evaluation results in Section~\ref{sec:evaluation}. After discussing related work in Section~\ref{sec:related}, we conclude in Section~\ref{sec:conclusion}.
\section{Background}
\label{sec:background}
In this section, we provide the necessary background on LiDAR object detection DNNs and anytime computing.


\subsection{LiDAR Object Detection DNNs}

The primary objective of LiDAR based object detection is to identify objects of interest within the detection area by processing input point clouds. Many LiDAR based object detection DNNs have been proposed~\cite{centerpoint,voxelnext,pointpillars}, some are optimized for latency, while others are optimized for accuracy. 

Figure~\ref{fig:baseline} illustrates the general workflow of LiDAR object detection DNNs. Their encoders are designed to extract features from the transformed input (e.g., voxels) with their backbone(s), typically by employing convolutional neural networks. An encoder can have a 3D backbone that applies sparse convolutions on 3D data, a 2D backbone similar to those used in vision object detection DNNs, or both. When both are used, the sparse output of the 3D backbone is projected to a bird-eye view (BEV) pseudo-image to turn it into a dense tensor so the 2D backbone can process it with dense convolutions.

After the encoder operation, the produced features are further processed by the decoder, which consists of one or more detection heads to output the 3D bounding boxes of the identified objects. When multiple detection heads are used, the targeted object classes are separated into groups depending on their size, and each detection head becomes responsible for one group~\cite{megvii}. Within each detection head, a series of convolutions is applied to infer various object attributes such as location, size, and velocity. Ultimately, non-maximum suppression or max pooling is used to extract the final results from predicted candidates.

\subsection{Sparse Convolution}
A point cloud $P$ is represented as an array of 3D point coordinates $(x, y, z)$, each accompanied by attributes such as LiDAR return intensity $i$.
\begin{equation}
    P = \{(x_1, y_1, z_1, i_1), \ldots, (x_n, y_n, z_n, i_n)\}
\end{equation}

Unlike 2D images, the indexes in the array of points do not inherently establish neighborhood relationships, creating a challenge for processing them with commonly used dense convolutional neural networks operating on dense tensors. To address this issue, point clouds are transformed into alternative representations, such as a 3D grid of fixed-size \textit{voxels} created by grouping spatially nearby points\cite{centerpoint,voxelnext}. These voxels can be represented as a 3D dense tensor and processed by 3D convolutions. However, this approach is avoided due to the significant computational overhead it incurs. Instead, voxels are represented as a sparse tensor and processed by sparse convolutions\cite{second}. A sparse tensor $V$ can be defined in coordinate list (COO) format, where each coordinate has a corresponding array of values. These values represent the features of each coordinate.


Sparse convolutions can yield the same result as dense convolutions while operating on sparse tensors. If the input tensor is significantly sparse, as in LiDAR point clouds, this saves a bulk of computational time compared to dense convolutions. For this reason, state-of-the-art LiDAR object detection DNNs commonly employ sparse convolutions.
Sparse convolutions apply given filters on all coordinates
where an input coordinate overlaps with any part of the filter. 

\begin{figure}[htp]
\centerline{\includegraphics[scale=0.15]{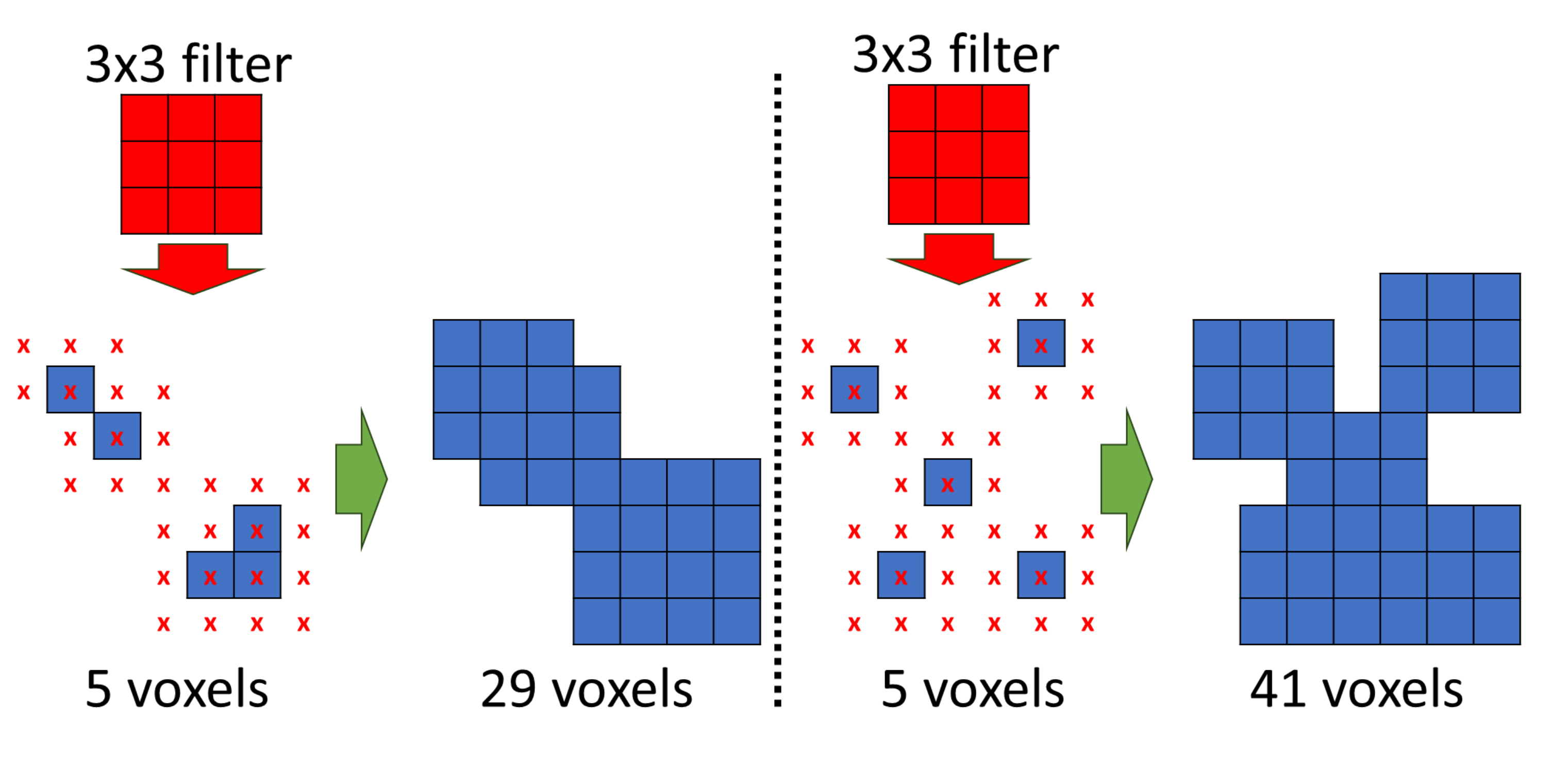}}
\caption{Two sparse convolution examples applying 3x3 filters. Blue squares indicate voxels. Red markings indicate the coordinates where the filter is applied.}
\label{fig:sparseconv}
\vspace{-15pt}
\end{figure}

It is important to note that a sparse convolution operation can generate a differently shaped output tensor, depending on the shape of the input tensor, as shown in Figure~\ref{fig:sparseconv}. As we will discuss in Section~\ref{sec:time-pred}, this introduces input-dependent timing variability in processing sparse convolutions. 

\subsection{Anytime Computing}
Anytime algorithms refer to a class of algorithms that can trade deliberation time for the quality of the results~\cite{boddy1989solving}. 
An anytime algorithm is capable of delivering a result whenever it is requested, and the quality of the result improves as the algorithm dedicates more time to finding the solution. 
For example, a path planning algorithm that progressively enhances its solution by continuously refining the path it has discovered can be considered as an anytime algorithm~\cite{boddy1989solving}.
In real-time systems, anytime algorithms are highly valuable for meeting dynamically changing deadlines as they can effectively trade-off between latency and quality.

Contract algorithms are a special type of anytime algorithms that require a predetermined time budget to be set prior to their activation~\cite{zilberstein1999contract}. They are non-interruptible and deliver results within the time budget, unlike arbitrarily interruptible anytime algorithms. In deadline-driven real-time systems, such as self-driving cars, contract algorithms can be used to effectively trade execution time for accuracy. Providing a framework to transform a LiDAR object detection DNN into a contract algorithm to make it deadline-aware is the primary focus of our work.

\section{Motivation}\label{sec:motivation}
To understand the requirements of an effective latency and accuracy trading approach, we profile two representative LiDAR object detection DNNs in detail on Jetson AGX Xavier.

Table~\ref{tbl:exectimetablepp} presents the execution time statistics for PointPillars~\cite{pointpillars}, a well-known LiDAR object detection DNN recognized for its low latency. We observe that approximately 79\% of the total processing time is consumed by its 2D backbone and detection heads. Therefore, a latency-accuracy tradeoff approach targeting these two stages can yield satisfactory results, as explored in a recent prior work\cite{alidar}.

However, when state-of-the-art LiDAR object detection DNNs are considered, an approach that only focuses on the 2D backbone and detection heads might not be efficient.

\begin{table}[htp]
\caption{Execution time (ms) statistics of PointPillars }
\label{tbl:exectimetablepp}
\begin{center}
\begin{tabular}{|l|c|c|c|c|}
\hline
Stage & Min & Average & 99th Perc. & Percentage  \\
\hline
Load to GPU  & 7.99  & 9.59 & 10.96 & 7\%\\
Feature Transform & 5.75  & 6.10  & 6.42  & 4\% \\
3D Backbone & 5.80  & 7.07  & 7.75  & 5\% \\
Project to BEV & 3.13  & 3.90  & 4.66  & 3\%\\
2D Backbone & 53.50 & 53.73 & 54.15 & 37\%\\
Detection Heads & 56.85 & 61.27 & 64.53 & 44\%\\
\hline
End-to-end  & 136.77 & 142.07 & 146.06 & 100\%  \\
\hline
\end{tabular}
\end{center}
\vspace{-5pt}
\end{table}

\begin{table}[htp]
\caption{Execution time (ms) statistics of CenterPoint.}
\begin{center}
\begin{tabular}{|l|c|c|c|c|}
\hline
Stage & Min & Average & 99th Perc. & Percentage  \\
\hline
Load to GPU  & 8.18 & 9.78 & 11.28 & 3\%\\
Feature Transform & 3.62  & 3.83 & 3.94  & 1\% \\
3D Backbone & 53.64  & 93.09  & 134.27  & 41\% \\
Project to BEV & 4.20  & 4.37  & 5.60  & 2\%\\
2D Backbone & 70.95 & 71.24 & 71.45 & 21\%\\
Detection Heads & 100.91 & 104.69 & 106.63 & 32\%\\
\hline
End-to-end  & 245.83 & 287.66 & 329.01 & 100\%  \\
\hline
\end{tabular}
\label{tbl:exectimetablecp}
\end{center}
\vspace{-15pt}
\end{table}

Table~\ref{tbl:exectimetablecp} shows the execution time breakdown of CenterPoint~\cite{centerpoint}, a recent 3D LiDAR object detection DNN that achieves higher detection accuracy than PointPillars~\cite{pointpillars}. Note that it spends significantly more time on the 3D backbone stage, accounting for 41\% of the total execution time. 


Although adopting sparse convolutions partially alleviates the computational burden of 3D backbone~\cite{second,torchsparse}, it still demands significant computational resources. Thus, the 3D backbone becomes another computational bottleneck, which must be addressed when trading accuracy for lower latency. 





One simple approach for achieving latency-accuracy tradeoff is training multiple models with varying input granularity (i.e., resolutions) and dynamically switching between them. However, this approach can be cumbersome during runtime due to the overhead involved in model switching (in terms of memory overhead and switching latency). It also necessitates training and fine-tuning a large number of models to achieve finely tuned tradeoffs.

Instead, we focus on developing a single model that can deliver the highest possible accuracy when there is flexibility with the deadline, while intelligently adjusting input data when the deadline becomes more stringent, as will be discussed in the next section.

\begin{figure*}[t]
\centerline{\includegraphics[scale=0.41]{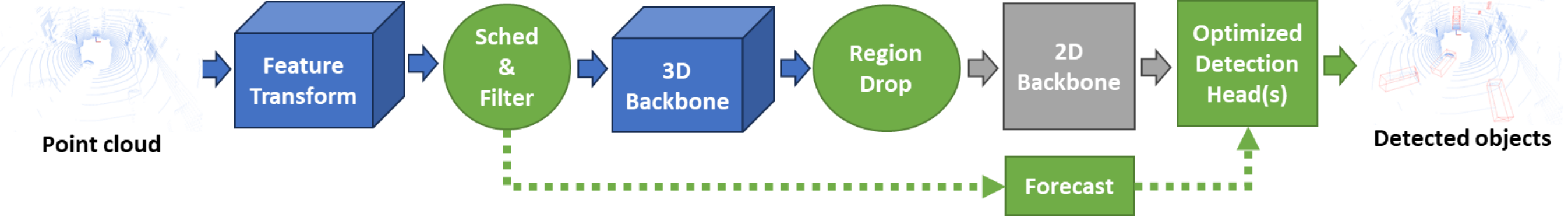}}
\caption{Overview of VALO.}
\label{proposed}
\vspace{-15pt}
\end{figure*}

\section{VALO}
\label{sec:VALO}

In this section, we introduce VALO, a scheduling framework that transforms a LiDAR object detection DNN into a non-interruptable anytime (contract) algorithm. VALO allows detection results to be produced in time for a gamut of deadline requirements, with a controlled tradeoff in accuracy.

\subsection{Overview}
The fundamental concept underpinning VALO's design is the \textit{scheduling of data} to facilitate tradeoffs between time and accuracy rather than scheduling the architectural components of the targeted DNN. This design choice makes VALO versatile, as it is not constrained by the architectural specifics of the LiDAR object detection DNNs.
Figure~\ref{proposed} illustrates VALO's three main components---scheduling, forecasting, and detection head optimization---highlighted in green, and their positions within the DNN pipeline. The region drop component is considered a part of scheduling.

First, VALO's scheduler comes into play after the DNN has completed the feature transformation stage. This allows it to make scheduling decisions at the voxel level instead of the raw point clouds, enabling more accurate predictions of the timing for the 3D backbone stage.

During the scheduling phase, VALO decides which regions of the input data will be processed to maximize detection accuracy within the deadline constraint. Once a decision is made, the data outside the selected regions is filtered out, and the remaining data is forwarded to the subsequent stage (Section~\ref{sec:region-sched}).

For effective region scheduling, VALO predicts execution times of subsequent network stages of each possible region selection (Section~\ref{sec:time-pred}). VALO also employs a mechanism to recover from execution time mispredictions  (Section~\ref{sec:region-drop}).

Next, while filtering part of the input can reduce latency, it can also negatively impact accuracy. To mitigate potential accuracy loss, VALO employs a forecasting mechanism that updates the positions of previously detected objects to the current time of execution. This operation is performed mostly in parallel while the DNN executes. After the detection heads generate object proposals, these proposals are combined with the list of forecasted objects. The combined list is then subjected to non-maximum suppression, which yields the final detection results (Section~\ref{sec:forecasting}).

Lastly, to further improve efficiency, we introduce a novel optimization technique for efficient detection head processing. This optimization technique eliminates the significant amount of redundant computation in detection heads without compromising detection accuracy (Section~\ref{sec:detheadopt}).

\subsection{Region Scheduling}
\label{sec:region-sched}
The scheduler decides which subset of input data (voxels) should be processed to meet a given deadline while maximizing accuracy. Intuitively, the less data it selects, the less time it takes for the DNN to process it, albeit at the expense of reduced accuracy. 
To make the scheduling problem tractable, we partition the fixed-size detection area into equally sized chunks along the $X$ (width) axis, which we refer to as \textit{regions}. 

\begin{figure}[htp] 
    \centering
    \begin{minipage}[b]{0.4\linewidth}
        \centering
        \includegraphics[width=\textwidth]{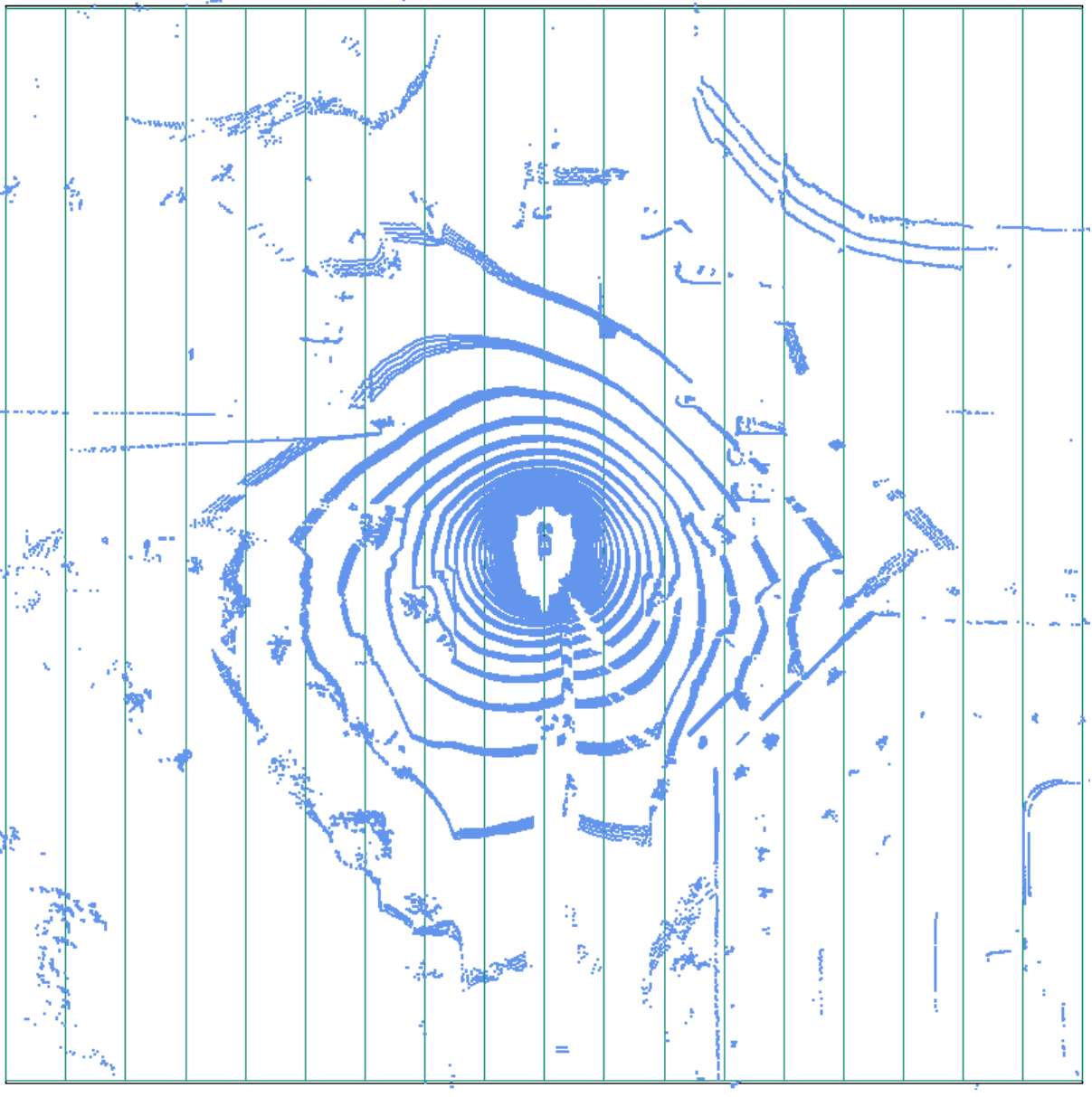}
        \subcaption{Partitioning example 1}
        \label{fig:partitioning_a}
    \end{minipage}
    \hspace{0.05\linewidth}
    \begin{minipage}[b]{0.4\linewidth}
        \centering
        \includegraphics[width=\textwidth]{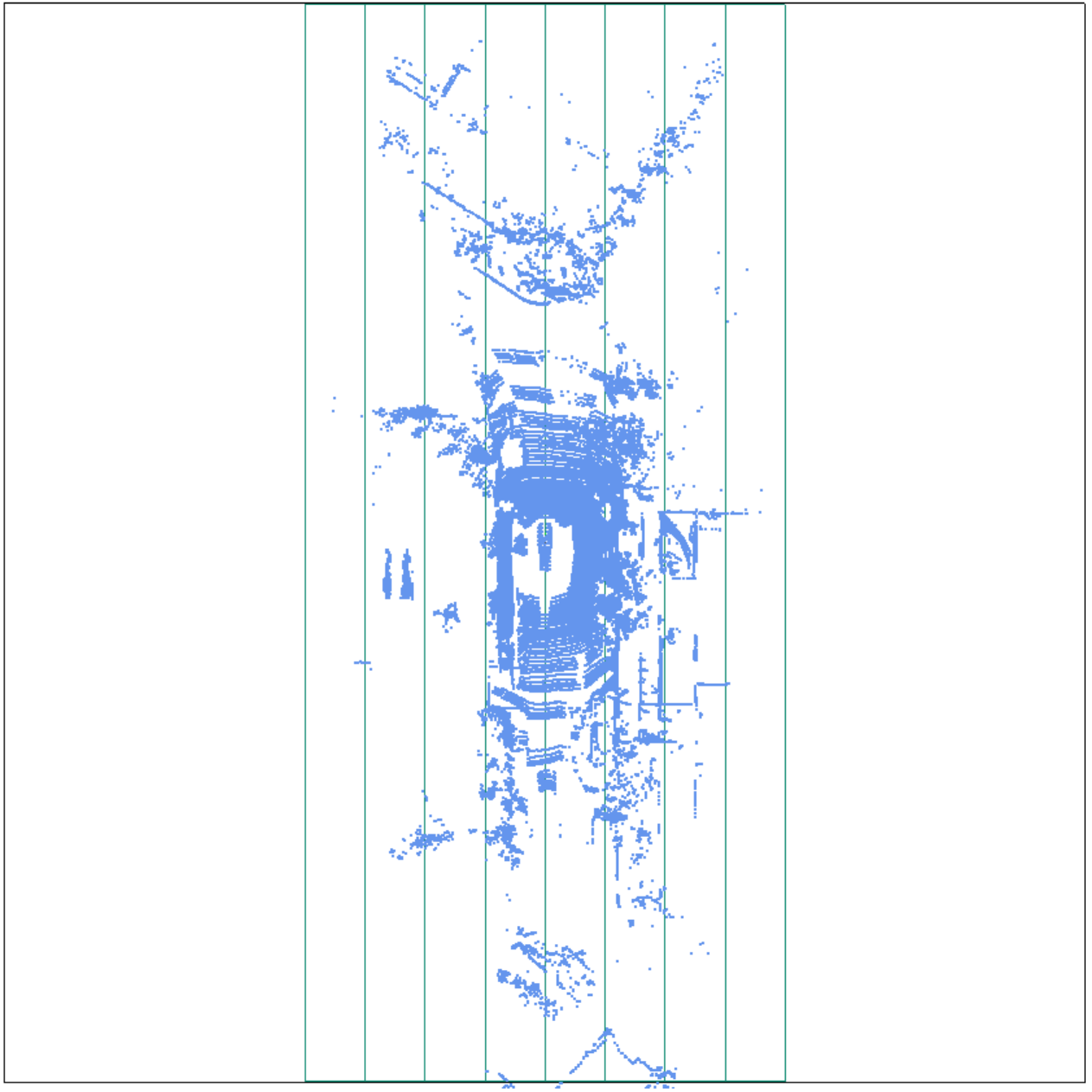}
        \subcaption{Partitioning example 2}
        \label{fig:partitioning_b}
    \end{minipage}
    \hfill
    \caption{Two examples of how the region scheduler partitions the detection area into regions.}
    \label{fig:partitioning}
\end{figure}

Figure~\ref{fig:partitioning} illustrates two examples of partitioning a $108 \times 108$ $m^2$ detection area into 18 vertical regions. In Figure~\ref{fig:partitioning_a}, the input point cloud is spread to all 18 regions. In contrast, Figure~\ref{fig:partitioning_b} shows that only a portion of the regions, 8 out of 18, contain points due to the structure of the environment scanned by LiDAR. 
In scenarios with empty regions, the scheduler skips all empty regions located before the first non-empty region and after the last non-empty region.
As a result, partitioning the input in the $X$ axis for some inputs allows for latency reduction without sacrificing accuracy in later stages. 

\begin{figure*}[t]
\centerline{\includegraphics[width=0.8\textwidth]{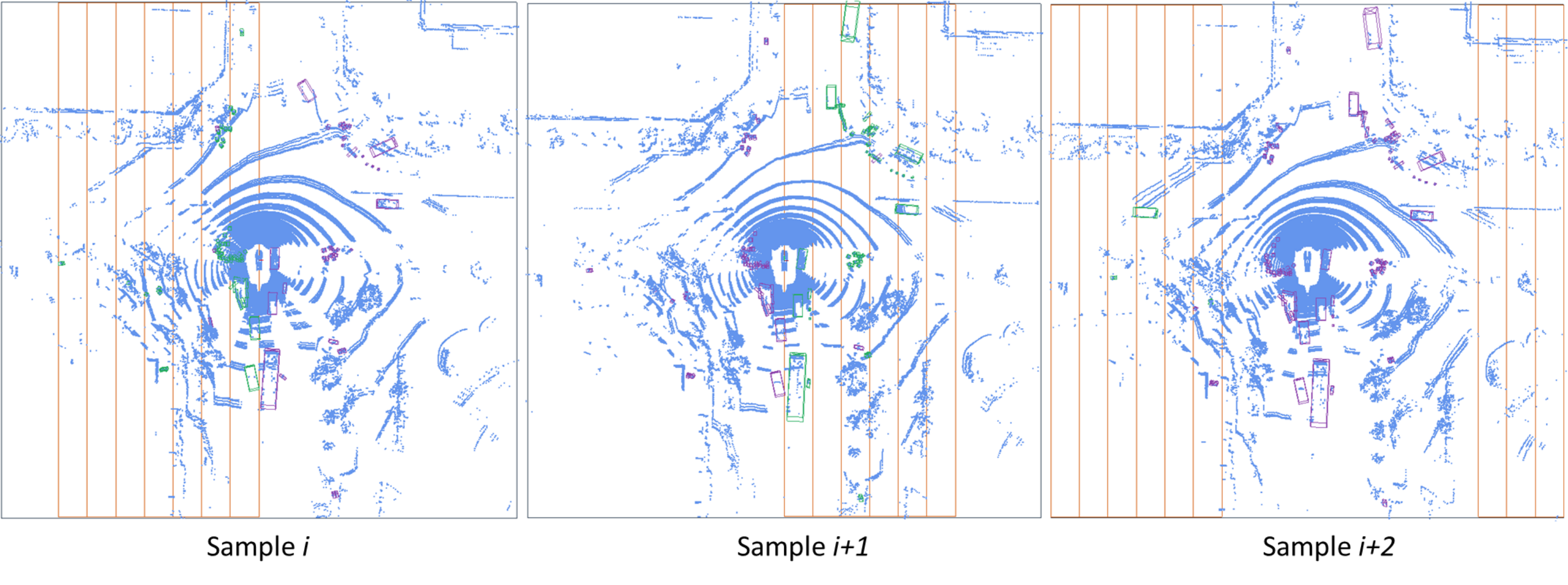}}
\caption{An example of region scheduling on three consecutive samples over time. The regions outlined in orange represent the selections made by the scheduler for processing. The green and purple bounding boxes indicate the objects detected as a result of processing the selected regions and the forecasted objects, respectively. Best viewed in color.}
\label{fig:sched_all}
\vspace{-10pt}
\end{figure*}

To determine which regions to process, we employ a greedy policy that sequentially selects the maximum number of input regions while adhering to the deadline constraint. Consequently, all regions are treated with equal priority. Figure~\ref{fig:sched_all} provides an illustrative example of the proposed region scheduling algorithm, which selects regions for processing over three consecutive inputs. For each input, the scheduler decides the regions to be scheduled for processing---starting from the next to the last of the previously scheduled regions---which can meet the given deadline.

Algorithm~\ref{alg:algo1} outlines our proposed scheduling algorithm. Initially, the scheduler counts the number of voxels in each region and returns the list of schedulable regions ($R_{S}$), and their voxel counts ($C_{S}$) (line 8).

The scheduler then reorders the obtained list so the selections start from the first non-empty region coming after $r_{last}$ (lines 9). Subsequently, candidate region selections are iterated from largest to smallest until one that meets the deadline is identified (lines 10-18).

\setlength{\textfloatsep}{2pt}
\begin{algorithm}[htp]
\DontPrintSemicolon
\textbf{Input}:\\
Input voxels ($V$),\\
Number of input regions ($N_R$),\\
Last scheduled region ($r_{last}$),\\
Relative deadline ($D$),\\
\textbf{Output}: Selected regions to be processed\\
\SetKwProg{funct}{function}{}{}
\SetKwFunction{func}{schedule}

\funct{\func{$V$, $N_R$, $r_{last}$, $D$}}
{
    $R_{S}, C_{S} \gets count\_voxels(V, N_R)$\\
    $R_{S}, C_{S} \gets reorder(R_{S}, C_{S}, r_{last})$

    $i \gets length\_of(R_{S})$\\
    \While{$i \geq 1$}
    {
        $R_{sel}, C_{sel} \gets R_{S}[:i], C_{S}[:i]$\\

        $E \gets calc\_wcet(R_{sel}, C_{sel})$\\
        $rem\_time \gets D - get\_elapsed\_time()$\\
        \If{$E < rem\_time$}
        {
            $i \gets 0$
        }
        \Else
        {
            $i \gets i-1$
        }
    }
}
\Return{$R_{sel}$}
\caption{Scheduling algorithm}
\label{alg:algo1}
\end{algorithm}

Once scheduling is completed, input voxels falling outside the selected regions ($R_{sel}$) are filtered, and the remaining voxels are forwarded to the 3D backbone as input. 
If the subsequent stage employs dense convolutions, the sparse output of the 3D backbone is then converted to a dense tensor where the regions are placed following the order in $R_{sel}$.

Our scheduling method brings three advantages. Firstly, selecting adjacent regions maintains spatial continuity and processes the input with minimal fragmentation, thereby avoiding accuracy degradation that can happen through slicing and batching nonadjacent regions. Secondly, it ensures a consistent level of ``freshness'' of object detection results over all regions, which is needed for effective forecasting operations (Section~\ref{sec:forecasting}). Thirdly, it incurs minimal scheduling overhead.

\subsection{Execution Time Prediction}
\label{sec:time-pred}
For effective region scheduling, 
the key \textit{challenge} is to determine whether a candidate list of regions can be processed within a given deadline constraint (Line 13 in Algorithm~\ref{alg:algo1}). The predicted execution time $E$ of a candidate list of regions can be calculated as:
\begin{equation}
    E = E_{S} +  E_{D} + E_{R}
\end{equation}
where 
$E_{S}$ is the time to process sparse data (i.e. 3D backbone), $E_{D}$ is the time to process dense data (i.e. 2D backbone and convolutions in detection heads), and $E_{R}$ is
the 
time to process the last stage of object detection task
such as non-maximum suppression.

For $E_{D}$, since the number of candidate regions ($|R_{sel}|$) determines the size of the dense input tensor that will be passed to the 2D backbone, it can be defined as a one-to-one function, where each possible $|R_{sel}|$ is mapped to an execution time determined through offline profiling. This mapping is feasible because the execution time of dense convolutions remains largely fixed as a function of input size, and there is a small finite number of possible regions.

\begin{figure}[htp]
\centerline{\includegraphics[width=0.45\textwidth]{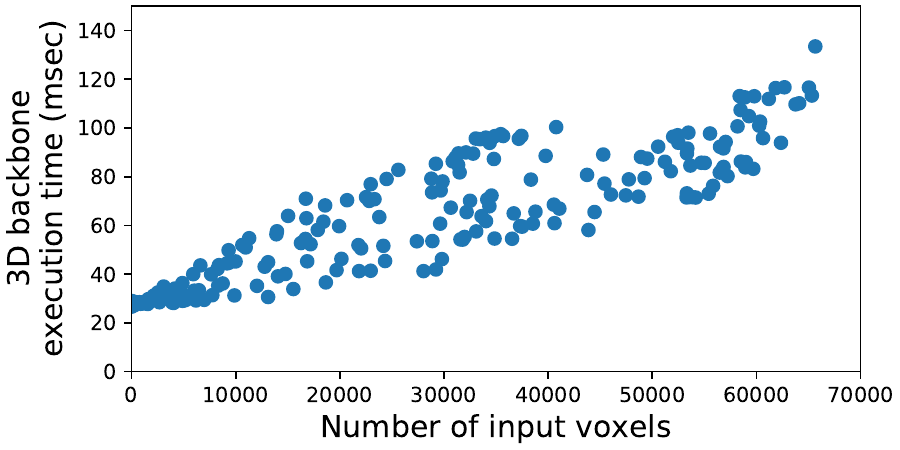}}
\caption{Profiled execution time CenterPoint's 3D Backbone.}
\label{fig:bb3dexectime}
\vspace{-5pt}
\end{figure}

On the other hand, $E_S$, the execution time of the sparse 3D backbone, is difficult to predict as it depends on the number of input voxels in a highly non-linear manner, as shown in Figure~\ref{fig:bb3dexectime}.

\begin{figure}[htp]
\centerline{\includegraphics[scale=0.30]{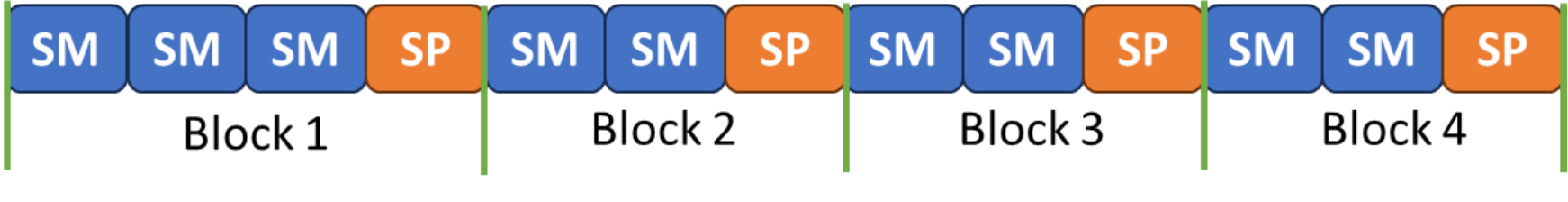}}
\caption{CenterPoint's 3D backbone broken into blocks. SM: Submanifold sparse convolution. SP: Sparse convolution.}
\label{fig:blocks}
\end{figure}

This non-linearity mainly stems from the fact that a sparse convolution layer can generate a different number of output voxels for the same number of input voxels depending on their relative positions, as illustrated in Figure~\ref{fig:sparseconv}. Consequently, the computational demand of processing a subsequent layer, which takes the output of the previous layer as input, will vary accordingly. 
To make the time prediction tractable, we break the 3D backbone into blocks at points where the count of forwarded voxels changes, as illustrated in Figure~\ref{fig:blocks}.

We then focus on separately predicting the execution time of each block. Note that, unlike a sparse convolution layer, batch normalization, activation functions, and submanifold sparse convolution~\cite{submanifold}, all of which heavily used in 3D backbones, do maintain the same input and output shapes (thus voxel counts), and thus can be safely grouped within a block. Denoting $V_i$ as the input voxels of a layer $L_i$, we define a block $B$ as:
\begin{equation}
    B = \{L_k, \ldots, L_l \mid \forall i,  \quad k \leq i \leq l, \quad |V_k| = |V_i|\}
\end{equation}
where $V_k$ is the input voxels of the first layer $L_k$. The input of a block $B$ denoted as $V_B$ is the same as $V_k$.

\begin{figure}[htp]
\centerline{\includegraphics[width=0.45\textwidth]{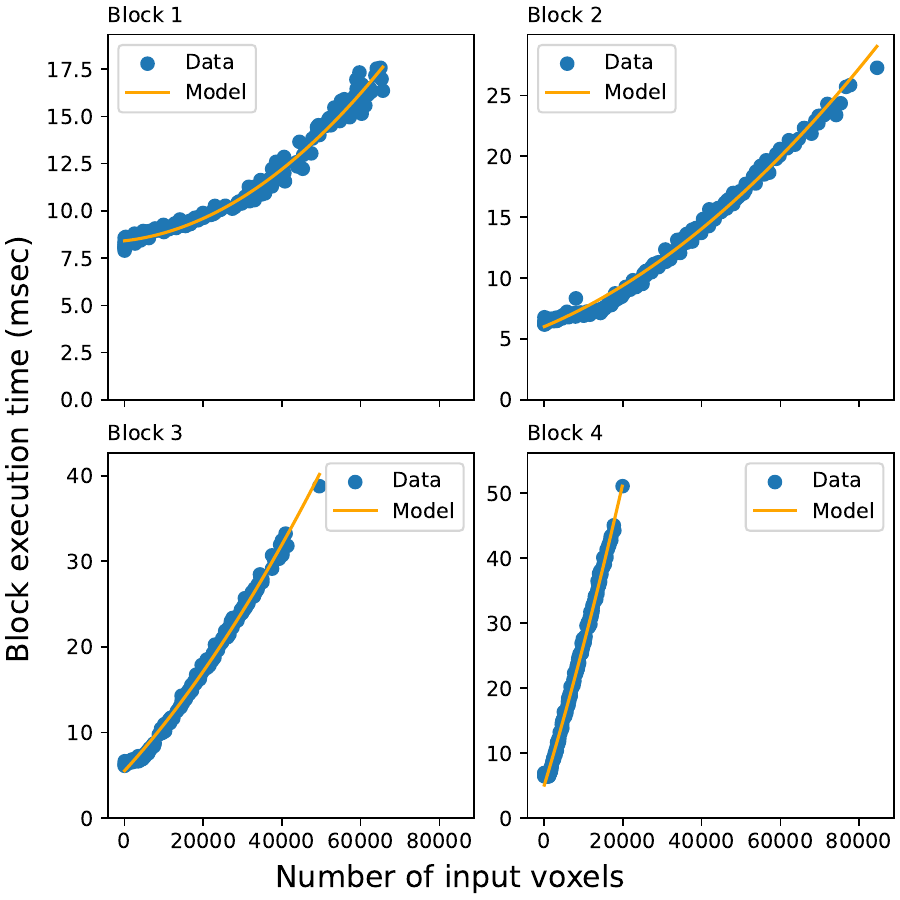}}
\caption{Profiled execution time of the blocks of CenterPoint's 3D backbone, and the quadratic models regressed from their execution times data.}
\label{fig:layerblockexectime}
\end{figure}

Figure~\ref{fig:layerblockexectime} shows the execution time profiles of all four blocks of the CenterPoint's 3D backbone. As can be seen in the figure, each block's execution time, as a function of the number of input voxels of the block, is more predictable using a simple quadratic prediction model. 

\begin{equation}
    \label{eq:quadratic}
    E_{B_i}(|V_{B_i}|) = \alpha |V_{B_i}|^2 + \beta |V_{B_i}| + \gamma
\end{equation}
where the coefficients $\alpha, \beta$, and $\gamma$ are determined by regression against the profiling data collected offline.
Then, the execution time of the 3D backbone can be predicted as follows:
\begin{equation}
    E_{S} = \sum_{i=1}^{n} E_{B_i}(|V_{B_i}|).
\end{equation}

However, a major challenge is that, except for the first block, the number of input voxels of the rest of the blocks, $C_{rest}$, are not known until the execution of the preceding blocks is completed.
\begin{equation}
    C_{rest} = \{|V_{B_2}|, \ldots, |V_{B_n}|\}
\end{equation}





To predict $C_{rest}$ for any given list of candidate input regions, we use a history-based approach, leveraging the fact that there is a strong similarity between consecutive LiDAR scans, as the movements of objects between scans are limited.
Specifically, for block $B_2$ to $B_n$, we keep track of each block's most recent input voxel counts of all input regions, which are updated whenever they are selected by the region scheduler and processed. Assuming voxel counts would be similar over time, we then aggregate the latest voxel counts of the current candidate regions to obtain $C_{rest}$.
Lastly, for $E_R$, the execution time to perform non-maximum suppression and other operations can vary depending on the number of object proposals in the detection pipeline. However, because it is relatively small compared to the rest of the pipeline, namely $E_D$ and $E_S$, we simply use the 99th percentile of the measured execution time through offline profiling, which provides a safe upper bound without significantly affecting time prediction accuracy. 

\subsection{Region Drop}
\label{sec:region-drop}
The aforementioned execution time prediction method for the 3D backbone can inevitably introduce some inaccuracy. 
For LiDAR object detection models with 2D backbones, such as CenterPoint~\cite{centerpoint}, 
after the execution of the 3D backbone, we additionally check if it will be possible to meet the deadline (see Figure~\ref{proposed}), considering the predicted execution time of the rest of the pipeline. If deemed not possible, we further reduce the number of input regions so that the deadline can be met. 
Note, however, that some recently proposed LiDAR object detection models such as VoxelNext~\cite{voxelnext} do not employ a 2D backbone as they are fully sparse. For such networks, the region dropping does not apply.

\subsection{Forecasting}
\label{sec:forecasting}

Forecasting estimates the present \textit{pose} of the objects identified in the past invocations of the object detector. Because our region scheduling method (Section~\ref{sec:region-sched}) can skip part of the input LiDAR scan due to deadline constraints, forecasting plays a critical role in mitigating the potential accuracy loss. 

We define a pose $P$ of an object at time $t$ as:
\begin{equation}
\label{eq:pose}
    P_t = \{T, S, \alpha, v, c, l\}
\end{equation}
where $T$ is the 3D coordinate of the object expressed in the LiDAR coordinate frame, $S$ is the 
bounding box, $\alpha$ is the heading angle, 
$v$ is the velocity vector,
$c$ is the confidence score,
and $l$ is the label (e.g., car or pedestrian). 
In this work, we focus on estimating $T$ and $\alpha$ and assume others to stay consistent over time.




The first part of forecasting involves maintaining a queue of previously detected object poses. 
For all processed input regions of an input frame, VALO
removes the old objects corresponding to processed regions from
the queue and appends the freshly detected objects in these
regions to the queue. Thus, the queue maintains the latest
detected objects of all regions.
 

The second part of forecasting involves performing mathematical calculations to estimate $P_{t_{cur}}$ for all objects in the pose queue. 
For each pose of an object in the pose queue, we first rotate and translate the object pose to be expressed in the global coordinate frame using the ego-vehicle pose. 
We then add the distance traveled by the object ($v \times (t_{cur} - t_{det})$) to the translation component ($T$) of the pose. 
Finally, we translate and rotate the pose to be expressed in the current LiDAR coordinate frame. 

At runtime, we update the queue on the CPU and perform actual pose updates on the GPU. 
We have developed a custom GPU kernel to update the poses of all objects in parallel. 
The forecasting GPU kernel is executed in a separate CUDA stream to maximize parallelism.


\subsection{Detection Head Optimization}
\label{sec:detheadopt}

LiDAR object detection DNNs include detection heads that are designed to extract specific attributes of objects, such as position, size, and orientation. Surprisingly, we discovered that a significant amount of redundant computations occur in processing the detection heads of state-of-the-art LiDAR object detection DNNs~\cite{openpcdet2020}. 


\begin{figure}[htp]
  \centering
  \includegraphics[width=0.4\textwidth]{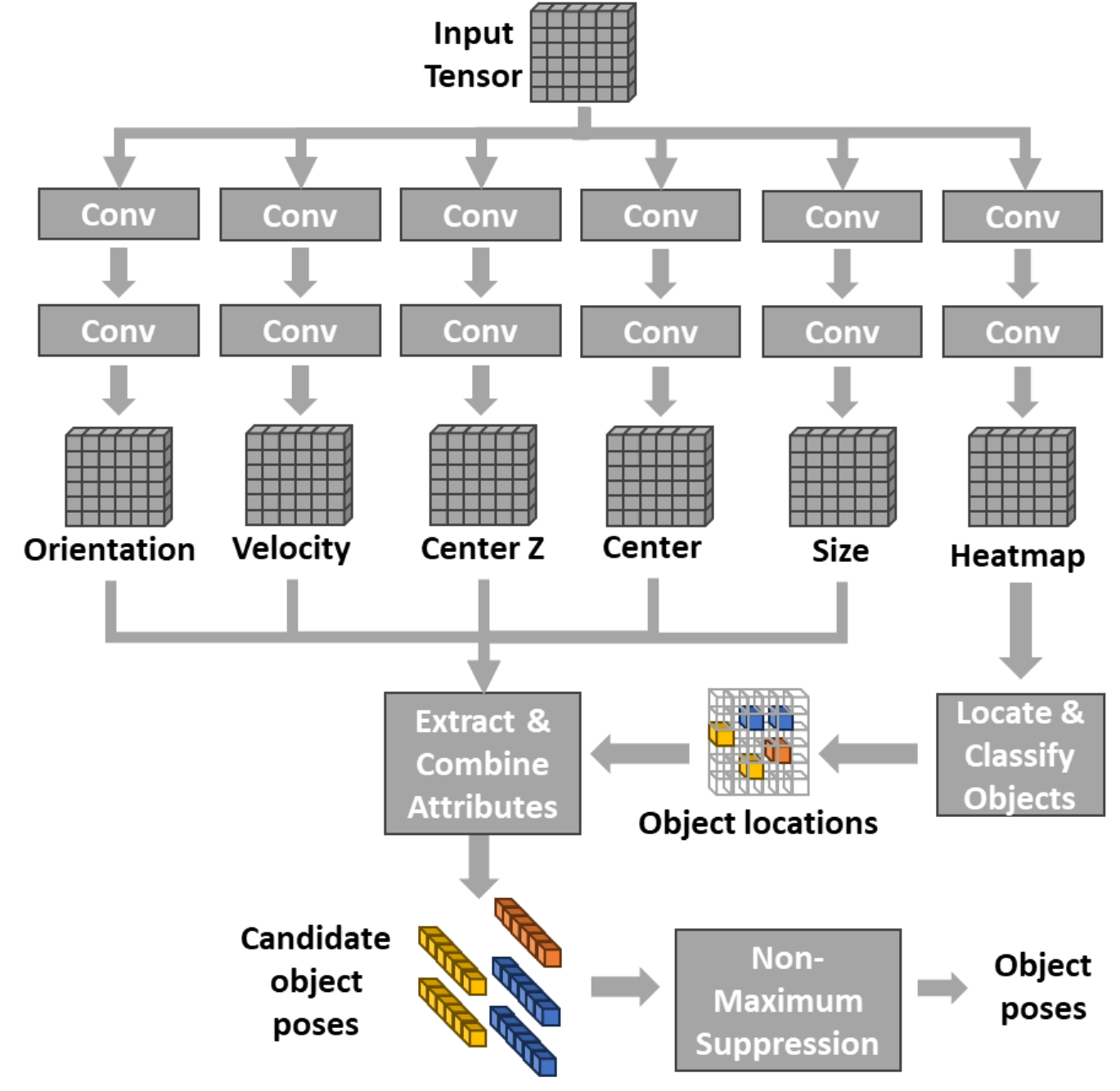} 
  \caption{General detection head architecture.}
  \label{fig:detheada}
\end{figure}

Figure~\ref{fig:detheada} illustrates the general architecture of a detection head, which performs a series of convolutions
to infer attributes of the objects. The width and height dimensions of the output tensors from these convolutions correspond to the width and height of the detection area in the bird's-eye view (BEV).
Among the inferred attributes, the \textit{heatmap} plays the most important role, as it holds the confidence scores of the objects used for classifying and locating them. In a
heatmap tensor, any score value above a predefined score threshold indicates an object proposal. The list of object proposals, $R$, extracted from the heatmap, can be expressed as:
\begin{equation}
    R = \{ (c_1, x_1, y_1), \ldots, (c_n, x_n, y_n) \}
\end{equation}
where $c$ is the confidence score and $x,y$ 
are a position in the detection area. Once $R$ is generated, remaining object attributes (e.g., orientation, velocity, size, etc.) are obtained from their corresponding output tensors at the $x, y$ positions in $R$, and combined into object poses (Eq.~\ref{eq:pose}). 

The problem with this approach is that it performs convolutions on all parts of the input while only the output locations that correspond to the object proposals ($R$) are utilized.
As a result, the convolutions inferring object attributes except the heatmap involve a significant amount of redundant computation.

To improve efficiency, we propose to optimize the detection head processing as follows:
    (1) First, the heatmap is computed in the same manner as in the baseline approach; 
    (2) Then the detected object list $R$ from the heatmap is utilized to selectively gather and batch small patches from the input tensor; 
    (3) Finally, convolutions are applied to this batch of patches to derive the object attributes.

\begin{figure}[htp]
  \centering
  \includegraphics[width=0.4\textwidth]{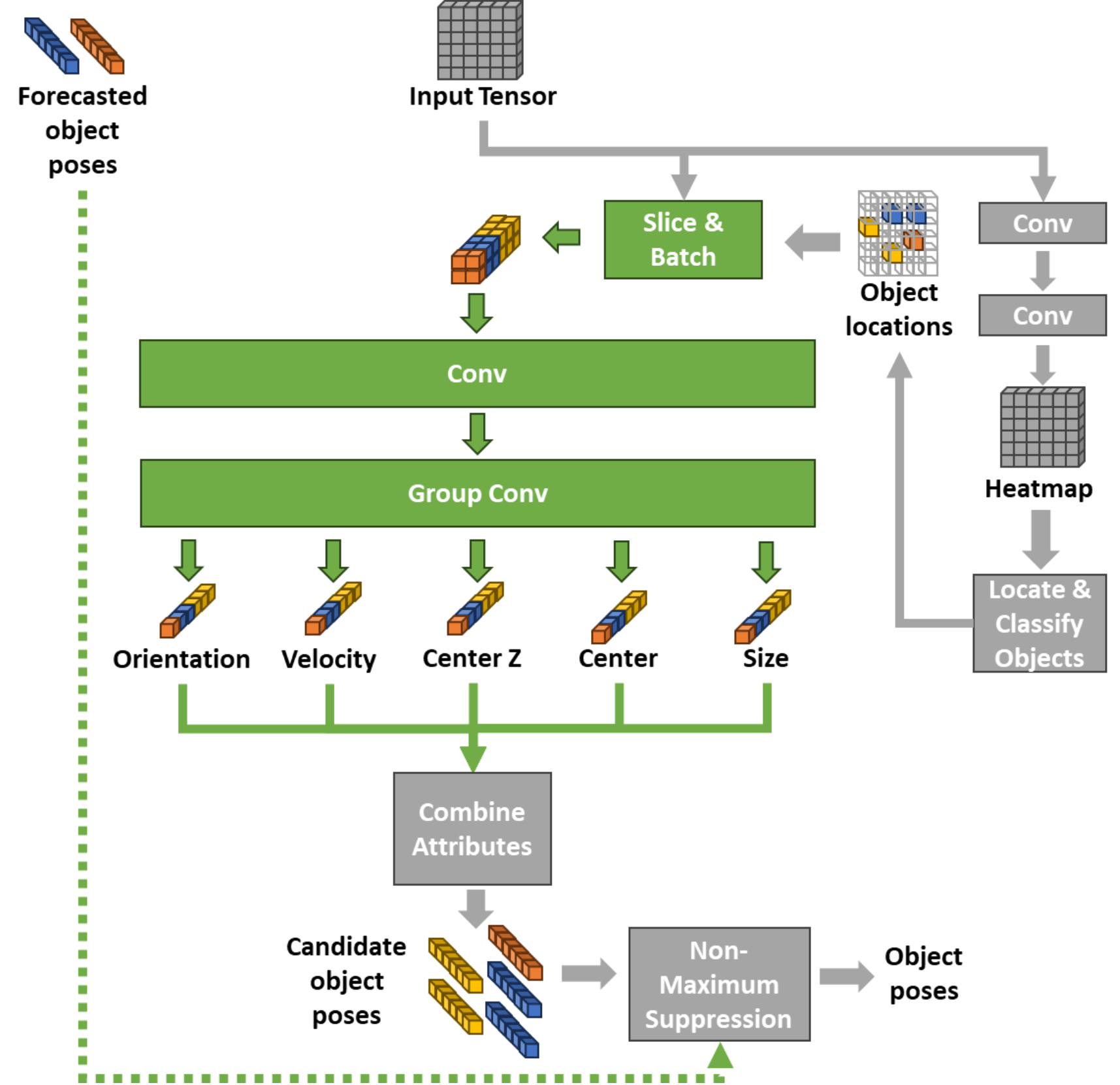} 
  \caption{Optimized detection head architecture.}
  \label{fig:detheadopt}
\end{figure}

Figure~\ref{fig:detheadopt} provides a visual representation of the proposed approach. 
Note that the proposed optimization ensures that convolutions are applied only to the data that is needed for producing the desired output corresponding to the locations in $R$. This approach significantly reduces the number of Multiply-Accumulate operations (MACs) without any loss of detection accuracy.

However, due to the reduction in the input size, there is a potential issue of GPU underutilization if we execute the attribute-inferring convolutions one by one as in the baseline.
To maximize GPU utilization, we concatenate them into a single convolution operation followed by a group convolution. This improves GPU utilization and reduces the GPU kernel invocation overhead. 

Note that some recent LiDAR object detection networks such as VoxelNext~\cite{voxelnext} employ sparse convolutions in detection heads instead of dense convolutions. For such a model, we replace the slice \& batch part of detection head optimization with filtering all sparse tensor coordinates that do not contribute to the output, and do not group the convolutions as they are sparse.
In this way, we significantly reduce computational overhead without losing detection accuracy and allow utilizing of the model trained for the baseline.
\section{Evaluation}\label{sec:evaluation}

For evaluation, we implemented VALO as an extension to OpenPCDet~\cite{openpcdet2020}, an open-source framework for LiDAR 3D object detection DNNs, which supports state-of-the-art methods. For this study, we mainly target CenterPoint~\cite{centerpoint} as a baseline and apply VALO to demonstrate its effectiveness. In addition, we also apply VALO on a more recently proposed VoxelNext~\cite{voxelnext}, a fully sparse DNN, to demonstrate the versatility of our approach. 


As for the dataset, we utilize nuScenes \cite{nuscenes}, a large-scale autonomous driving dataset, and use nuScenes detection score (NDS)\cite{nuscenes} as the detection accuracy metric since it was reported to correlate with the driving performance better than the classic average precision (AP) metric~\cite{offlineeval2023}. 
In the rest of the evaluation, unless noted otherwise, we normalize the NDS score with respect to the maximum NDS score we observed among the all compared methods.
We utilize 30 distinct scenes from the nuScenes evaluation dataset, with each scene containing annotated LiDAR scans spanning 20 seconds, sampled at intervals of 350 milliseconds. The sample period is chosen to match the worst-case execution time of the slowest baseline method 
on our evaluation platform.






To capture the timeliness aspect of the detection performance, we evaluated the methods under a range of deadline constraints, from 350 to 90 milliseconds. 
The deadline range is chosen to be between 
the best-case execution time of the fastest baseline method and the worst-case execution time of the slowest baseline model. 
During each test, we kept a buffer holding the latest detection results and updated this buffer every time the method being tested met the deadline. In case of a deadline miss, we considered the buffered detection results as the output and ignored the produced ones by assuming the job was aborted. 

As for the hardware platform, we used an NVIDIA Jetson AGX Xavier\cite{jetson-agx}, equipped with 16 GiBs of RAM, for runtime performance evaluation. We maximized all hardware clocks and allocated the GPU resources only for the method being tested. For software, we used Jetson JetPack 5.1 and Ubuntu 20.04. Training of the models was done on a separate desktop machine with an NVIDIA RTX 4090 GPU.

We present the evaluation results in the following three subsections. First, we compare VALO with a set of baselines to evaluate its performance. Second, we perform an ablation study to demonstrate the benefits of VALO's components. Lastly, we shift our focus to the intrinsic details of VALO and analyze the execution time behavior of its components.

\subsection{Comparison With the Baselines}


Below is the list of methods we compared in this section:
\begin{itemize}
\item \textbf{CenterPoint}\cite{centerpoint}: This is a representative state-of-the-art LiDAR object detection network architecture that employs a voxel encoder as its 3D backbone, followed by a region-proposal-based 2D backbone and six detection heads, each of which focuses on a subset of the object classes \cite{megvii}. Before being forwarded to the 3D backbone, the input point cloud is transformed into fixed-sized voxels. The size of a voxel is a design parameter of the network, which should stay consistent during training and testing. In this work, we consider three voxel configurations $75 \times 75 \times 200$ $mm^3$, $100 \times 100 \times 200$ $mm^3$, and $200 \times 200 \times 200$ $mm^3$, which are called CenterPoint75, CenterPoint100, and CenterPoint200, respectively. Employing bigger voxels reduces the computing cost at the expense of accuracy.

\item \textbf{VoxelNext}\cite{voxelnext}: A recently proposed LiDAR object detection network, featuring a voxel encoder as its 3D backbone deeper than CenterPoint's followed by six detection heads. Unlike CenterPoint, all convolutions in its detection heads operate on sparse tensors.
Like CenterPoint, VoxelNext also can be configured to have a different voxel size. We focus on only the setting that employs voxels of size $75 \times 75 \times 200$ $mm^3$ (i.e. VoxelNext75). 

\item \textbf{AnytimeLidar}\cite{alidar}: To the best of our knowledge, this is the only work that can provide runtime latency and accuracy trade-off (i.e. anytime computing) for LiDAR object detection DNNs in literature. It achieves the anytime capability by utilizing early exits in processing the 2D backbone and skipping a subset of detection heads dynamically. 
While AnytimeLidar is originally based on PointPillars~\cite{pointpillars}, we ported it to the CenterPoint75 baseline to make a fair comparison, which we call AnytimeLidar-CP75. Note that AnytimeLidar cannot be applied to VoxelNext since it lacks a 2D backbone.

\item \textbf{VALO}: The proposed method in this work. VALO can be applied to CenterPoint and VoxelNext baselines. We call VALO-CP75 and VALO-VN75 when it is applied to the CenterPoint75 and VoxelNext75 baselines, respectively.
\end{itemize}

\subsubsection{VALO vs. AnytimeLidar}

In this experiment, we compare the performance of VALO and AnytimeLidar with the CenterPoint75 baseline from which they are applied. 



\begin{figure}[htp]  
    \centering

    \begin{subfigure}[b]{0.45\textwidth}  
        \includegraphics[width=\textwidth]{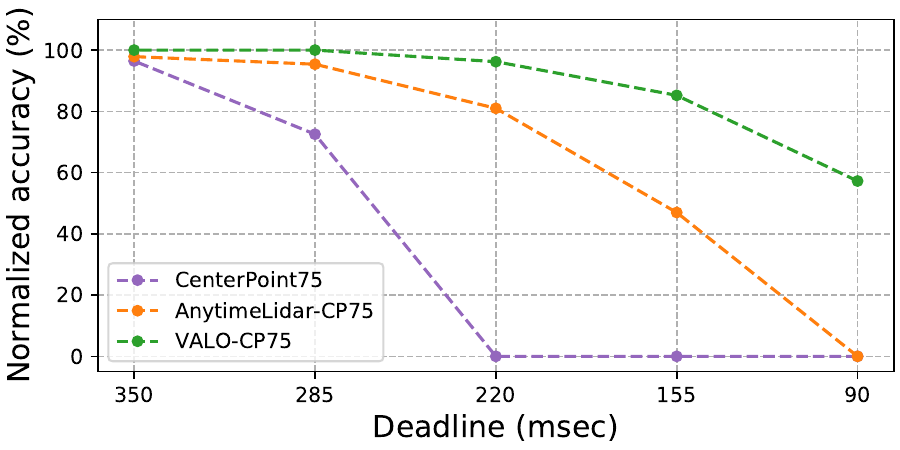}
        \caption{Detection accuracies.}
        \label{fig:acc1}
    \end{subfigure}
    \hfill  
    \begin{subfigure}[b]{0.45\textwidth}
        \includegraphics[width=\textwidth]{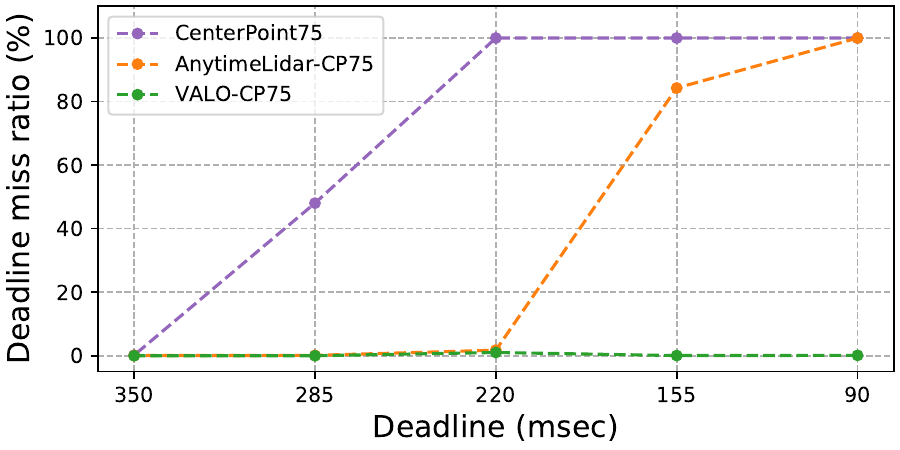}
        \caption{Deadline miss rates.}
        \label{fig:dl1}
    \end{subfigure}

    \caption{VALO vs. AnytimeLidar on CenterPoint}
    \label{fig:combined-anytime}
\end{figure}

Figure~\ref{fig:combined-anytime} shows the results. Figure~\ref{fig:acc1} compare how detection accuracy changes in relation to the varying deadline constraints. Figure~\ref{fig:dl1}, on the other hand, compare the corresponding deadline miss rates of the tested methods under the deadline constraints.

Note first that, under the 350 milliseconds deadline constraint, all methods can meet the deadline without a need for tradeoffs and demonstrate their maximum accuracy.
When the deadline tightens, however, the CenterPoint baseline immediately begins to miss deadlines as it cannot adjust its computing demand according to the given deadline, resulting in a significant drop in accuracy. AnytimeLidar and VALO, on the other hand, can trade accuracy for lower latency (i.e., anytime capable), and thus achieve improved performance as they can meet the deadlines better. However, when the deadline is 155 milliseconds, AnytimeLidar starts to miss deadlines due to its limited anytime computing capability. But VALO respects the deadline constraints down to 90 milliseconds and achieves higher accuracy.

AnytimeLidar falls short of matching the effectiveness of VALO primarily due to dismissing the contribution of the 3D backbone on the total latency. 
Moreover, AnytimeLidar's effectiveness will be further reduced if a single detection head architecture, instead of multi-head detection architecture in this work, is used because its ability to make a trade-off is in large part enabled by skipping a subset of the detection heads, which is possible only in multi-head architecture.

In contrast, VALO can make fine-grained execution time and accuracy tradeoffs, primarily due to its ability to schedule a portion of data to process, independent of the neural network architectural specifics, such as 3D/2D backbone or the number of detection heads. This distinct focus on data makes VALO a more versatile framework that can be applied in any LiDAR object detection DNN. 

\subsubsection{VALO vs. other non-anytime baselines}

\begin{figure}[htp]
\centerline{\includegraphics[width=0.45\textwidth]{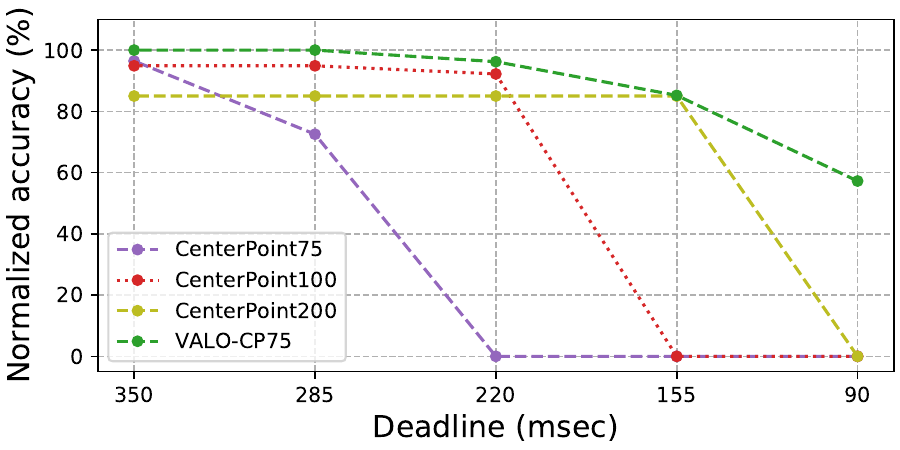}}
\caption{VALO vs. CenterPoint variants.}
\label{fig:acc2}
\end{figure}

Figure~\ref{fig:acc2} shows the detection performance of VALO-CP75 and three other CenterPoint baselines. All baselines have distinct execution time demands and accuracy they can deliver. For example, when the deadline is 350ms, CenterPoint75 achieves the best accuracy among the three baselines. But when the deadline is 220ms, CenterPoint75's accuracy falls down to zero because it no longer is able to meet the deadline. On the other hand, CenterPoint200's accuracy does not change all the way down to the deadline of 155ms as it can still meet the deadline albeit at a somewhat lower accuracy. Note, however, that these baseline models are fixed and cannot make accuracy vs. latency trade-offs on the fly at runtime. 
VALO, on the other hand, can adapt itself to a wide range of deadline constraints---from 90 to 350ms---on the fly while providing the best possible accuracy for a given deadline constraint.

As an alternative way to adapt to the varying deadline constraints on the fly, one can consider using multiple DNN models of differing latency-accuracy tradeoffs (like CenterPoint75, 100, and 200 in this experiment) and switch between them depending on a given deadline constraint at runtime as done in~\cite{d3}. However, the problems of such an approach are that it needs to train, fine-tune, and manage all these models separately. Furthermore, these models need to be loaded into the precious (GPU) memory all the time for real-time operations, even when only one of them is actually used at a time. In contrast, VALO can make such trade-offs at runtime from a single model without requiring any additional memory overhead.


\subsubsection{VALO on VoxelNext}

To demonstrate VALO's versatility, we applied it to VoxelNext~\cite{voxelnext}, which has a significantly different architecture than CenterPoint. Unlike CenterPoint, VoxelNext does not use a 2D backbone and instead relies solely on 3D sparse convolution layers.

\begin{figure}[htp]
\centerline{\includegraphics[width=0.45\textwidth]{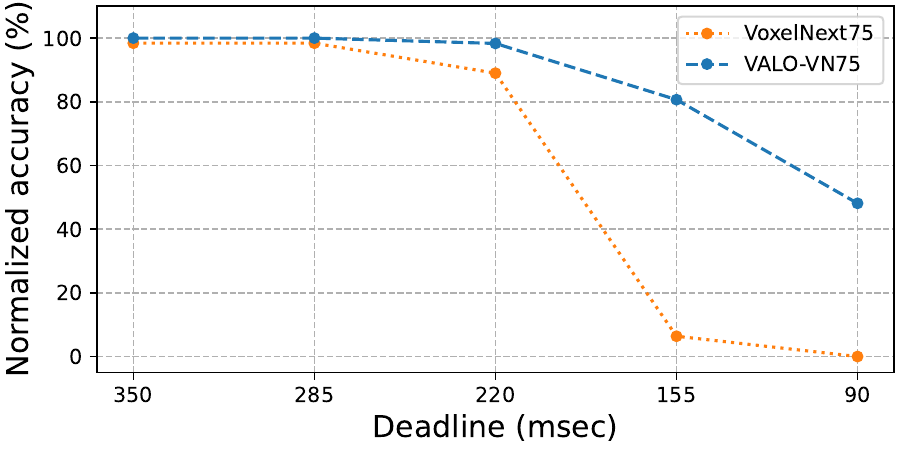}}
\caption{VALO on VoxelNext.}
\label{fig:acc3}
\vspace{-5pt}
\end{figure}

Figure~\ref{fig:acc3} shows the result. As in the CenterPoint case, VALO-VN75 performs better than the baselines in all deadline constraints. The region scheduling (Section~\ref{sec:region-sched}) allows VALO-VN75 to dynamically adjust the time spent on the 3D backbone and the detection heads effectively, effectively making it anytime capable. 

\subsubsection{Effectiveness of Time Prediction}

The effectiveness of VALO's region scheduling critically depends on the accuracy of its time prediction (Section~\ref{sec:time-pred}). To evaluate the effectiveness of the proposed history-based time prediction method, we compare its accuracy with a simple quadratic prediction model that directly predicts the execution time of the entire 3D backbone from the number of input voxels (as opposed to predicting per-block based prediction in our proposed history-based time prediction approach). We denote this baseline method as \textit{quadratic} whereas our history-based approach as \textit{history}.

\begin{figure}[htp]
\centerline{\includegraphics[width=0.45\textwidth]{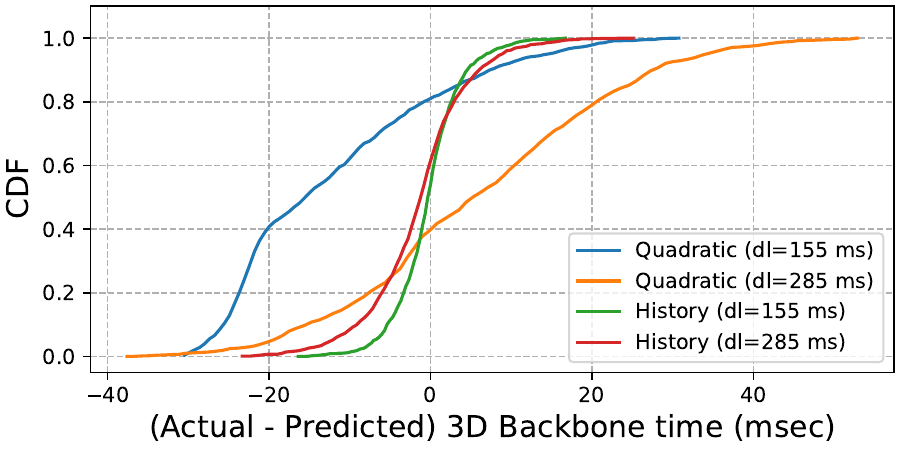}}
\caption{Cumulative distribution function of time prediction error for history-based and baseline methods.}
\label{fig:timeprederr}
\vspace{-5pt}
\end{figure}

Figure~\ref{fig:timeprederr} compares the accuracy of both time prediction methods in predicting 3D backbone execution time against the evaluation dataset. As can be seen in the figure, our history-based prediction method significantly outperforms the baseline quadratic method, which helps reduce deadline violations and improve detection accuracy. 

\subsection{Ablation Study}
\label{sec:ablationstudy}
In this experiment, we investigate the contribution of region scheduling and forecasting by comparing VALO with its two variants explained below. We also include the CenterPoint75 baseline for comparison.

\begin{itemize}
\item \textbf{VALO-NSNF-CP75}: This variant of VALO operates without scheduling (Section~\ref{sec:region-sched}) and forecasting (Section~\ref{sec:forecasting}), hence denoted as `No Scheduling No Forecasting' (NSNF). 
However, it does perform 
detection head optimization (Section~\ref{sec:detheadopt}).
\item \textbf{VALO-NF-CP75}: This variant of VALO performs region scheduling (Section~\ref{sec:region-sched}) and detection head optimization (Section~\ref{sec:detheadopt}), but not forecasting (Section~\ref{sec:forecasting}).
\end{itemize}

\begin{figure}[htp]
\centerline{\includegraphics[width=0.45\textwidth]{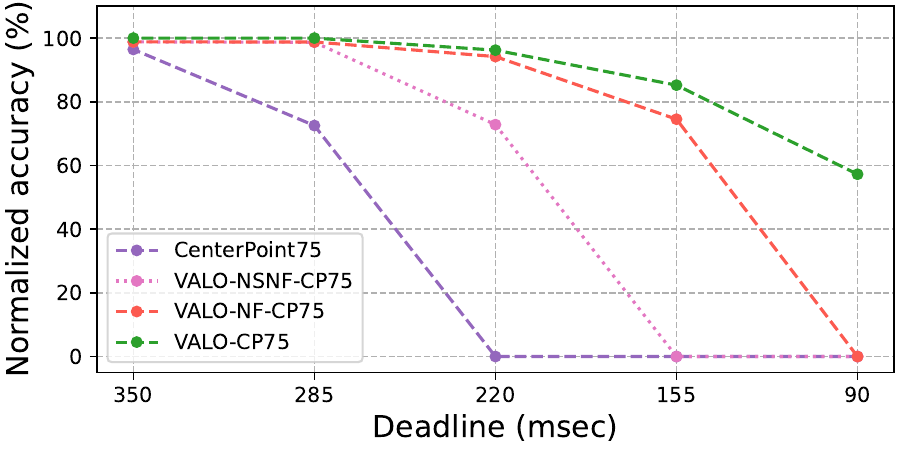}}
\caption{Detection accuracy achieved by the variants of VALO.}
\label{fig:acc4}
\vspace{-5pt}
\end{figure}

Figure~\ref{fig:acc4} presents the experimental results where we observe improved performance as additional VALO components are introduced to the baseline CenterPoint75. 
First, VALO-NSNF-CP75 achieves a higher accuracy over the baseline CenterPoint75 when the deadline is tighter than 350 milliseconds. For instance, at the 285 milliseconds deadline, VALO-NSNF-CP75 matches the accuracy of CenterPoint75 at 350 milliseconds. This underscores the effectiveness of
detection head optimization in reducing execution time without compromising accuracy. 
Next, VALO-NF further improves accuracy across a wider range of deadline constraints by enabling region scheduling because it can make execution time and accuracy tradeoffs, preventing deadline misses and boosting accuracy over VALO-NSNF. 
Lastly, VALO achieves the highest accuracy across all deadline constraints by additionally utilizing forecasting, 
which is particularly effective on tight deadlines. 
This is because forecasting plays a more crucial role when the number of scheduled regions reduces as the deadline tightens. 

\subsection{Component-level Timing Analysis}
\label{sec:component_analysis}

In this experiment, we delve into the execution timing characteristics of the components of VALO when it is applied to the CenterPoint75. 

\begin{figure*}[htp]  
    \centering

    \begin{subfigure}[b]{0.3\linewidth} 
        \includegraphics[width=\textwidth]{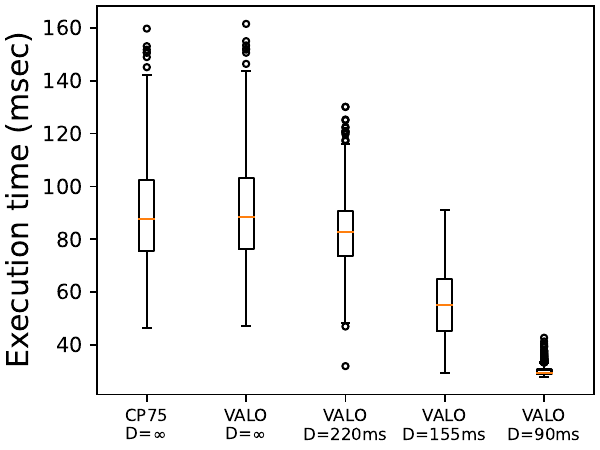}
        \caption{3D backbone (Voxel encoder)}
        \label{fig:bb3dbp}
    \end{subfigure}
    \hfill
    \begin{subfigure}[b]{0.3\linewidth}
        \includegraphics[width=\textwidth]{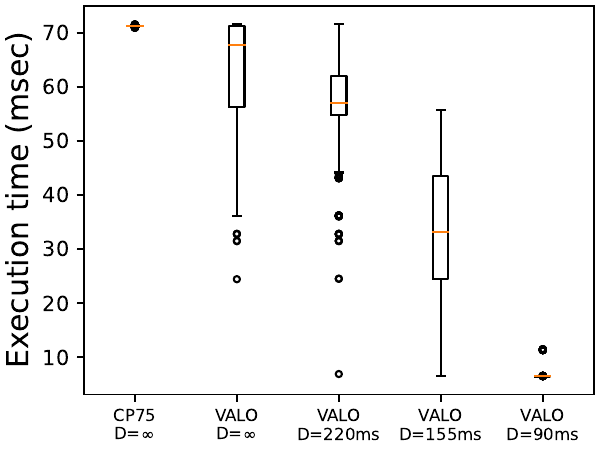}
        \caption{2D backbone (RPN)}
        \label{fig:bb2dbp}
    \end{subfigure}
    \hfill
    \begin{subfigure}[b]{0.3\linewidth}
        \includegraphics[width=\textwidth]{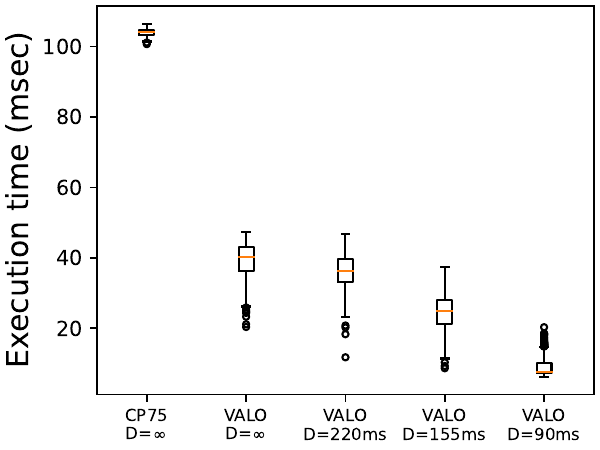}
        \caption{Detection head (CenterHead)}
        \label{fig:detheadbp}
    \end{subfigure}
    \caption{Component-level execution time profile of the baseline and VALO on Centerpoint75 under different deadline constraints}
    \label{fig:component-combined}
    \vspace{-10pt}
\end{figure*}

Figure~\ref{fig:component-combined} shows the execution timing of the 3D backbone, 2D backbone, and detection heads. For each component, we consider five different cases. The first two involve using CenterPoint75 and VALO-CP75, where there is no deadline. The rest are the results of VALO-CP75 executed with 220, 115, and 90-millisecond deadline constraints, respectively.

\subsubsection{3D Backbone}
Figure~\ref{fig:bb3dbp} shows the execution time profile of the 3D backbone portion of the network. 
Note first that the CenterPoint75 baseline shows a high degree of variations, influenced by the varying count and positioning of input voxels. 
When there is no deadline, the time spent on the 3D backbone of VALO-CP75 is about the same as CenterPoint75 as expected.
As the deadline gets tighter, however, 
VALO's execution time of the 3D backbone is progressively reduced because its region scheduler dynamically selects a subset of input regions that can be executed within the given time budget.

\subsubsection{2D Backbone}

Figure~\ref{fig:bb2dbp} shows the execution time profile of processing the 2D backbone, where convolutions on dense tensors take place. 
Unlike the 3D backbone processing, even when there is no deadline, we can observe a notable decrease in the execution time in VALO compared to the CenterPoint75 baseline. This is because our data partitioning scheme (Section~\ref{sec:region-sched}), which exploits the sparsity of LiDAR data, can skip empty input regions in the 2D backbone, thus reducing latency. As the deadline get tighter, we also observe a further reduction in the execution time of the 2D backbone as a result of reduced input data selected by the scheduler.

\subsubsection{Detection Heads}

Figure~\ref{fig:detheadbp} shows the execution time profile of processing the detection head. 
Note first that we observe more than 50\% reduction in detection head processing latency on VALO-CP75 compared to the CenterPoint75 baseline even when there is no deadline constraint. This is due to the proposed detection head optimization described in Section~\ref{sec:detheadopt}, which significantly reduce the amount of data to be processed by eliminating redundant data. 
In addition, as the deadline get tighter, we again observe progressive reduction in execution time in VALO due to further reduction in the input data to the detection head thanks to its scheduler. 


\subsubsection{Overhead}
\label{sec:overhead}
We measured 3 milliseconds of scheduling overhead in the worst case, including the input filtering time. There is also 3 milliseconds overhead due to the voxel counting operations as a part of history-based time prediction. We did not observe any overhead incurred by the forecasting operation when end-to-end latency is considered, as it is efficiently executed in parallel with the backbones. 
Note that the total overhead of VALO on CenterPoint75 is only about 6 milliseconds, which is less than 2\% of the average execution time of CenterPoint75.

\section{Related Work}
\label{sec:related}

Timely execution of autonomous driving software is essential to ensure safe and efficient navigation. Traditionally, the timing requirements (i.e. deadlines) of the autonomous driving tasks are often fixed at the design time~\cite{autoware, apollo}, which is not adaptable to the highly varying execution time demands~\cite{alcon2020timing}. Recently, Gog et al.\cite{d3} have highlighted the potential benefits of adopting a flexible approach, which can dynamically change deadlines in the autonomous driving software based on the specific driving situation, such as the speed of the vehicle or sudden pedestrian appearance, to improve performance and safety of the vehicle. 

LiDAR object detection is a critical component in many autonomous driving systems~\cite{lidaro}. With the release of large-scale autonomous driving datasets~\cite{nuscenes,sun2020waymo}, researchers have developed deep learning-based object detection models that achieve state-of-the-art performance. Besides aiming to achieve high accuracy, recent work has also considered reducing latency as an objective~\cite{pointpillars,pillarnet,shi2020parta2,voxelnext,centerpoint,ada3d,liu2022spatial} for real-time operation. These works can achieve remarkable accuracy in real-time when executed on high-end GPUs and accelerators. However, their deployment on edge computing platforms such as Jetson AGX Xavier~\cite{jetson-agx} still poses a challenge due to their significant computational overhead and latency. More importantly, they lack the capability to dynamically adapt their execution time in a deadline-aware manner, which is needed for real-time cyber-physical systems. 



Recent studies have explored the concept of ``anytime perception" for neural networks, which enables them to execute within defined deadlines while making tradeoffs between execution time and accuracy. For example, Kim et al.\cite{anytimenet} achieved this by iteratively adding layers to an image classification network and retraining it to incorporate ``early exits." Lee et al.\cite{subflow} focused on the neuron level, prioritizing critical neurons for accuracy while deactivating others to save time. Bateni et al.\cite{apnet} used per-layer approximation instead of early exits and presented a scheduling solution for multiple deep neural network tasks. Yao et al.\cite{imprecisecomp} also dealt with the scheduling of multiple deep neural network tasks, utilizing imprecise computation alongside early exits. 
While these works primarily targeted image classification tasks, object detection tasks present unique challenges. 

Heo et al.\cite{multipath} introduced a multi-path deep neural network architecture designed for anytime perception in vision-based object detection. Another work by the same authors~\cite{rtscale} designed an adaptive image scaling method that respects the deadline constraints for the multi-camera object detection task. Gog et al.\cite{pylot} proposed to switch between DNNs to make latency and accuracy tradeoffs dynamically at runtime. 
Hu et al.\cite{hu2021exploring} suggested reducing the resolution of less critical parts of the scene to lower computational costs. Lie et al.~\cite{rttasksched, liu2020removing} divided individual image frames into smaller sub-regions with varying levels of criticality, using LiDAR data to batch-process essential sub-regions to meet deadlines. However, these prior efforts mainly focus on 2D vision and do not account for the unique characteristics of 3D point cloud processing. 

Recently, Soyyigit et al.\cite{alidar} proposed a set of techniques that enable anytime capability for LiDAR object detection DNNs. They focused on object detection models where the bulk of the computation is performed on the 2D backbone and detection heads, such as PointPillar~\cite{pointpillars} and Pillarnet~\cite{pillarnet}. However, the effectiveness of their approach diminishes on recent state-of-the-art object detection models where the bulk of time is spent on 3D backbone~\cite{centerpoint, voxelnext}.
Fundamentally, such effort that focuses on model-level improvements may fail to work when the architecture of the model changes. 
In contrast, our work focuses on data-level scheduling, independent of the architectural details of the backbones and detection heads, and thus can be seamlessly applied to any state-of-the-art LiDAR object detection DNNs.

\section{Conclusion}\label{sec:conclusion}

In this work, we presented VALO, a versatile anytime computing framework for LiDAR object detection DNNs.
VALO's superior performance compared to the prior state-of-the-art 
comes from three major contributions: (i) partitioning the input data into regions and efficiently scheduling them with the goal of maximizing accuracy while respecting the deadlines, (ii) lightweight forecasting of the previously detected objects to mitigate potential accuracy loss due to partially processing the input, (iii) and intelligently reducing redundant computations in processing the detection heads of the object detection neural network with no loss of accuracy.
Evaluation results have shown that our approach can adapt to a wide-range of deadline constraints in processing 
LiDAR object detection DNNs, and enables a fine-grained and effective execution time and accuracy tradeoff.


\section*{Acknowledgments}\label{sec:acknowledge}
This research is supported in part by NSF grants CNS-1815959, CPS-2038923, III-2107200, and CPS-2038658.

\bibliography{references}

\begin{thebibliography}{10}

\bibitem{centerpoint}
T.~Yin, X.~Zhou, and P.~Krähenbühl, ``Center-based 3d object detection and tracking,'' in {\em IEEE/CVF Conference on Computer Vision and Pattern Recognition (CVPR)}, 2021.

\bibitem{pointpillars}
A.~H. Lang, S.~Vora, H.~Caesar, L.~Zhou, J.~Yang, and O.~Beijbom, ``Pointpillars: Fast encoders for object detection from point clouds,'' in {\em IEEE/CVF Conference on Computer Vision and Pattern Recognition (CVPR)}, 2019.

\bibitem{voxelnext}
Y.~Chen, J.~Liu, X.~Zhang, X.~Qi, and J.~Jia, ``Voxelnext: Fully sparse voxelnet for 3d object detection and tracking,'' in {\em IEEE/CVF Conference on Computer Vision and Pattern Recognition (CVPR)}, 2023.

\bibitem{d3}
I.~Gog, S.~Kalra, P.~Schafhalter, J.~E. Gonzalez, and I.~Stoica, ``D3: a dynamic deadline-driven approach for building autonomous vehicles,'' in {\em European Conference on Computer Systems (EuroSys)}, 2022.

\bibitem{li2023red}
Z.~Li, T.~Ren, X.~He, and C.~Liu, ``Red: A systematic real-time scheduling approach for robotic environmental dynamics,'' in {\em IEEE Real-Time Systems Symposium (RTSS)}, 2023.

\bibitem{anytimenet}
J.-E. Kim, R.~Bradford, and Z.~Shao, ``Anytimenet: Controlling time-quality tradeoffs in deep neural network architectures,'' in {\em Design, Automation Test in Europe Conference Exhibition (DATE)}, 2020.

\bibitem{apnet}
S.~Bateni and C.~Liu, ``Apnet: Approximation-aware real-time neural network,'' in {\em IEEE Real-Time Systems Symposium (RTSS)}, 2018.

\bibitem{imprecisecomp}
S.~Yao, Y.~Hao, Y.~Zhao, H.~Shao, D.~Liu, S.~Liu, T.~Wang, J.~Li, and T.~Abdelzaher, ``Scheduling real-time deep learning services as imprecise computations,'' in {\em IEEE International Conference on Embedded and Real-Time Computing Systems and Applications (RTCSA)}, 2020.

\bibitem{rttasksched}
S.~Liu, S.~Yao, X.~Fu, H.~Shao, R.~Tabish, S.~Yu, A.~Bansal, H.~Yun, L.~Sha, and T.~Abdelzaher, ``Real-time task scheduling for machine perception in intelligent cyber-physical systems,'' {\em IEEE Transactions on Computers}, vol.~71, no.~8, 2022.

\bibitem{abc}
J.-E. Kim, R.~Bradford, M.-K. Yoon, and Z.~Shao, ``Abc: Abstract prediction before concreteness,'' in {\em Design, Automation Test in Europe Conference Exhibition (DATE)}, 2020.

\bibitem{alidar}
A.~Soyyigit, S.~Yao, and H.~Yun, ``Anytime-lidar: Deadline-aware 3d object detection,'' in {\em IEEE International Conference on Embedded and Real-Time Computing Systems and Applications (RTCSA)}, 2022.

\bibitem{nuscenes}
H.~Caesar, V.~Bankiti, A.~H. Lang, S.~Vora, V.~E. Liong, Q.~Xu, A.~Krishnan, Y.~Pan, G.~Baldan, and O.~Beijbom, ``nuscenes: A multimodal dataset for autonomous driving,'' in {\em IEEE/CVF Conference on Computer Vision and Pattern Recognition (CVPR)}, 2020.

\bibitem{jetson-agx}
{NVIDIA}, ``{Jetson AGX Xavier Developer Kit}.'' developer.nvidia.com/embedded/jetson-agx-xavier-developer-kit.

\bibitem{megvii}
B.~{Zhu}, Z.~{Jiang}, X.~{Zhou}, Z.~{Li}, and G.~{Yu}, ``{Class-balanced Grouping and Sampling for Point Cloud 3D Object Detection},'' {\em arXiv preprint arXiv:1908.09492}, 2019.

\bibitem{second}
Y.~Yan, Y.~Mao, and B.~Li, ``Second: Sparsely embedded convolutional detection,'' {\em Sensors}, vol.~18, no.~10, 2018.

\bibitem{boddy1989solving}
M.~Boddy and T.~L. Dean, {\em Solving time-dependent planning problems}.
\newblock Brown University, Department of Computer Science, 1989.

\bibitem{zilberstein1999contract}
S.~Zilberstein, F.~Charpillet, and P.~Chassaing, ``Real-time problem-solving with contract algorithms,'' in {\em International Joint Conference on Artificial Intelligence (IJCAI)}, 1999.

\bibitem{torchsparse}
H.~Tang, Z.~Liu, X.~Li, Y.~Lin, and S.~Han, ``Torchsparse: Efficient point cloud inference engine,'' in {\em Conference on Machine Learning and Systems (MLSys)}, 2022.

\bibitem{submanifold}
B.~Graham and L.~van~der Maaten, ``Submanifold sparse convolutional networks,'' {\em arXiv preprint arXiv:1706.01307}, 2017.

\bibitem{openpcdet2020}
O.~D. Team, ``Openpcdet: An open-source toolbox for 3d object detection from point clouds.'' https://github.com/open-mmlab/OpenPCDet, 2020.

\bibitem{offlineeval2023}
T.~Schreier, K.~Renz, A.~Geiger, and K.~Chitta, ``On offline evaluation of 3d object detection for autonomous driving,'' in {\em IEEE/CVF International Conference on Computer Vision Workshops (ICCVW)}, 2023.

\bibitem{autoware}
S.~Kato, S.~Tokunaga, Y.~Maruyama, S.~Maeda, M.~Hirabayashi, Y.~Kitsukawa, A.~Monrroy, T.~Ando, Y.~Fujii, and T.~Azumi, ``Autoware on board: Enabling autonomous vehicles with embedded systems,'' in {\em ACM/IEEE International Conference on Cyber-Physical Systems (ICCPS)}, 2018.

\bibitem{apollo}
``{Baidu Apollo team (2017), Apollo: Open Source Autonomous Driving}.'' https://github.com/ApolloAuto/apollo.

\bibitem{alcon2020timing}
M.~Alcon, H.~Tabani, L.~Kosmidis, E.~Mezzetti, J.~Abella, and F.~J. Cazorla, ``Timing of autonomous driving software: Problem analysis and prospects for future solutions,'' in {\em IEEE Real-Time and Embedded Technology and Applications Symposium (RTAS)}, 2020.

\bibitem{lidaro}
Y.~Li and J.~Ibanez-Guzman, ``Lidar for autonomous driving: The principles, challenges, and trends for automotive lidar and perception systems,'' {\em IEEE Signal Processing Magazine}, vol.~37, no.~4, 2020.

\bibitem{sun2020waymo}
P.~Sun, H.~Kretzschmar, X.~Dotiwalla, A.~Chouard, V.~Patnaik, P.~Tsui, J.~Guo, Y.~Zhou, Y.~Chai, B.~Caine, {\em et~al.}, ``Scalability in perception for autonomous driving: Waymo open dataset,'' in {\em IEEE/CVF Conference on Computer Vision and Pattern Recognition (CVPR)}, 2020.

\bibitem{pillarnet}
C.~M. Guangsheng~Shi, Ruifeng~Li, ``Pillarnet: Real-time and high-performance pillar-based 3d object detection,'' in {\em European Conference on Computer Vision (ECCV)}, 2022.

\bibitem{shi2020parta2}
S.~Shi, Z.~Wang, J.~Shi, X.~Wang, and H.~Li, ``From points to parts: 3d object detection from point cloud with part-aware and part-aggregation network,'' {\em IEEE Transactions on Pattern Analysis and Machine Intelligence}, vol.~43, no.~8, 2020.

\bibitem{ada3d}
T.~Zhao, X.~Ning, K.~Hong, Z.~Qiu, P.~Lu, Y.~Zhao, L.~Zhang, L.~Zhou, G.~Dai, H.~Yang, and Y.~Wang, ``Ada3d : Exploiting the spatial redundancy with adaptive inference for efficient 3d object detection,'' in {\em IEEE/CVF International Conference on Computer Vision (ICCV)}, 2023.

\bibitem{liu2022spatial}
J.~Liu, Y.~Chen, X.~Ye, Z.~Tian, X.~Tan, and X.~QI, ``Spatial pruned sparse convolution for efficient 3d object detection,'' in {\em Advances in Neural Information Processing Systems (NeurIPS)}, 2022.

\bibitem{subflow}
S.~Lee and S.~Nirjon, ``Subflow: A dynamic induced-subgraph strategy toward real-time dnn inference and training,'' in {\em IEEE Real-Time and Embedded Technology and Applications Symposium (RTAS)}, 2020.

\bibitem{multipath}
S.~Heo, S.~Cho, Y.~Kim, and H.~Kim, ``Real-time object detection system with multi-path neural networks,'' in {\em IEEE Real-Time and Embedded Technology and Applications Symposium (RTAS)}, 2020.

\bibitem{rtscale}
S.~Heo, S.~Jeong, and H.~Kim, ``Rtscale: Sensitivity-aware adaptive image scaling for real-time object detection,'' in {\em Euromicro Conference on Real-Time Systems (ECRTS)}, 2022.

\bibitem{pylot}
I.~Gog, S.~Kalra, P.~Schafhalter, M.~A. Wright, J.~E. Gonzalez, and I.~Stoica, ``Pylot: A modular platform for exploring latency-accuracy tradeoffs in autonomous vehicles,'' in {\em IEEE International Conference on Robotics and Automation (ICRA)}, 2021.

\bibitem{hu2021exploring}
Y.~Hu, S.~Liu, T.~Abdelzaher, M.~Wigness, and P.~David, ``On exploring image resizing for optimizing criticality-based machine perception,'' in {\em IEEE International Conference on Embedded and Real-Time Computing Systems and Applications (RTCSA)}, 2021.

\bibitem{liu2020removing}
S.~Liu, S.~Yao, X.~Fu, R.~Tabish, S.~Yu, A.~Bansal, H.~Yun, L.~Sha, and T.~Abdelzaher, ``On removing algorithmic priority inversion from mission-critical machine inference pipelines,'' in {\em IEEE Real-Time Systems Symposium (RTSS)}, 2020.

\end{thebibliography}
\clearpage


\end{document}